\documentclass{article}

\usepackage{microtype}
\usepackage{graphicx}
\usepackage{subfigure}
\usepackage{booktabs} % for professional tables
\usepackage{textcomp}  
\usepackage{booktabs}

\usepackage{hyperref}
\usepackage{paralist}

\usepackage[accepted]{icml2025}

\usepackage{amsmath}
\usepackage{amssymb}
\usepackage{mathtools}
\usepackage{amsthm}
\usepackage[utf8]{inputenc}
\usepackage{array}
\usepackage{xcolor}
\usepackage[capitalize,noabbrev]{cleveref}

\theoremstyle{plain}

\theoremstyle{definition}

\theoremstyle{remark}

\usepackage[textsize=tiny]{todonotes}

\icmltitlerunning{CAD-Editor}

\begin{document}

\twocolumn[
\icmltitle{CAD-Editor: A Locate-then-Infill Framework with \\ Automated Training Data Synthesis for Text-Based CAD Editing}




\icmlsetsymbol{work}{$^\dagger$}

\begin{icmlauthorlist}
\icmlauthor{Yu Yuan}{ustc,work}
\icmlauthor{Shizhao Sun}{msra}
\icmlauthor{Qi Liu}{ustc}
\icmlauthor{Jiang Bian}{msra}
\end{icmlauthorlist}

\icmlaffiliation{ustc}{University of Science and Technology of China}
\icmlaffiliation{msra}{Microsoft Research Asia}

\icmlcorrespondingauthor{Shizhao Sun}{shizsu@microsoft.com}

\icmlkeywords{Machine Learning, Computer-Aided Design}

\vskip 0.3in
]




\printAffiliationsAndNotice{{$^\dagger$ Work done during the internship at Microsoft Research Asia. Open-source research project starts from March 2024.}
{Yu Yuan: \textless yyhappier@mail.ustc.edu.cn\textgreater ,}
{Qi Liu: \textless qiliuql@ustc.edu.cn\textgreater, }
{Jiang Bian: \textless jiabia@microsoft.com\textgreater.}}

 % otherwise use the standard text.

\begin{figure*}[h]
\begin{center}
\includegraphics[width=\textwidth]{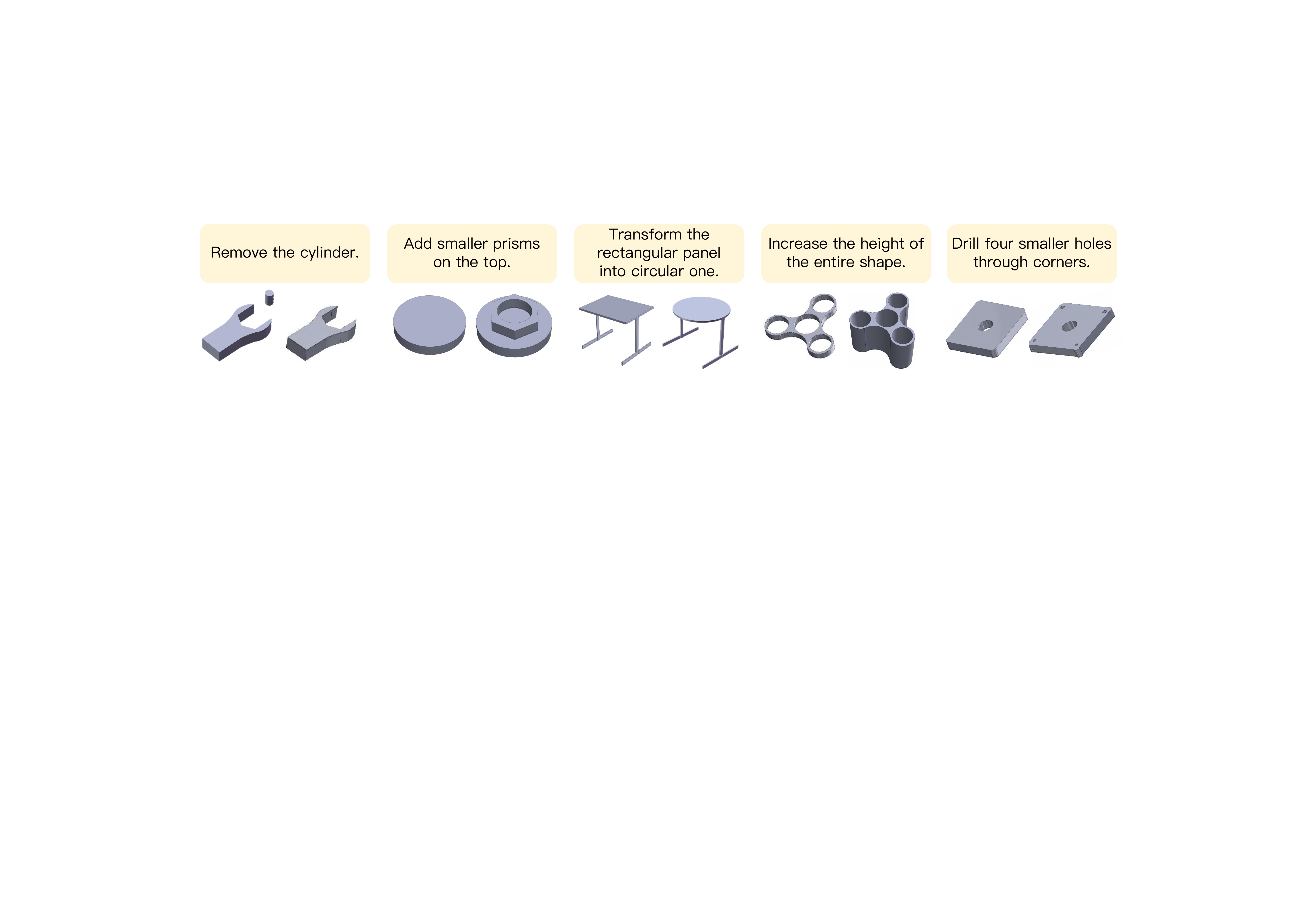}  
\end{center}
\caption{Text-based CAD editing achieved by CAD-Editor. Each sub-figure shows the editing instruction at the top, the original CAD model on the left, and the edited CAD model on the right. The rendered image is shown for better comprehension. The actual editing occurs on sketch-and-extrusion (SE) operations of a CAD model to provide editability and reusability.}
\label{fig:teaser}
\end{figure*}

\begin{abstract}
Computer Aided Design (CAD) is indispensable across various industries. 
\emph{Text-based CAD editing}, which automates the modification of CAD models based on textual instructions, holds great potential but remains underexplored.
Existing methods primarily focus on design variation generation or text-based CAD generation, either lacking support for text-based control or neglecting existing CAD models as constraints.
We introduce \emph{CAD-Editor}, the first framework for text-based CAD editing. 
To address the challenge of demanding triplet data with accurate correspondence for training, we propose an automated data synthesis pipeline. 
This pipeline utilizes design variation models to generate pairs of original and edited CAD models and employs Large Vision-Language Models (LVLMs) to summarize their differences into editing instructions.
To tackle the composite nature of text-based CAD editing, we propose a locate-then-infill framework that decomposes the task into two focused sub-tasks: locating regions requiring modification and infilling these regions with appropriate edits. 
Large Language Models (LLMs) serve as the backbone for both sub-tasks, leveraging their capabilities in natural language understanding and CAD knowledge.
Experiments show that CAD-Editor achieves superior performance both quantitatively and qualitatively. The code is available at \url {https://github.com/microsoft/CAD-Editor}.
\end{abstract}

\section{Introduction}
\label{introduction}

In the modern digital era, Computer-Aided Design (CAD) has become indispensable across industries. 
Most modern CAD tools follow the \textbf{Sketch-and-Extrude (SE) Operations} paradigm~\citep{shahin2008feature, camba2016parametric}, where designers sketch 2D curves to define the outer and inner boundaries of profiles, extrude them into 3D shapes, and combine these shapes to create complex models.

CAD model creation is an iterative process, where an initial draft undergoes multiple modifications until it aligns with user requirements. 
Natural language plays a crucial role throughout this process, serving as a key medium of communication.
For non-experts, it offers the most intuitive way to express their needs, while for professionals, it enables fast, detailed, and precise instructions.
Consequently, a system capable of automatically editing CAD models based on textual instructions -- known as \textbf{text-based CAD editing} (Figure~\ref{fig:teaser}) -- has the potential to revolutionize the entire CAD design workflow. 
Such a system could significantly accelerate the development of CAD models and empower a broader range of individuals, especially those with limited design expertise, to create CAD models more effectively.

While important, text-based CAD editing receives limited attention. 
Some studies explore \emph{design variation generation}, where new CAD models are generated by randomly altering components of an existing model~\citep{wu2021deepcad, xu2022skexgen, xu2023hierarchical, zhang2024flexcad}. 
However, these approaches lack support for text-based control over the appearance of the generated CAD models, limiting their practical usability.
Another line of research makes initial attempts at \emph{text-based CAD generation}, focusing on generating new CAD models directly from textual descriptions~\citep{khantext2cad, li2024cad}. 
Nonetheless, these methods do not incorporate an existing CAD model as input, which prevents them from leveraging the original design's context and constraints.

Text-based CAD editing presents several distinct challenges.
First, training for this task requires \emph{triplet data with accurate correspondence} among an original CAD model, an editing instruction, and an edited CAD model.
However, such data does not naturally exist, and manually collection is both costly and difficult to scale.
Second, text-based CAD editing is inherently a \emph{composite problem}. 
It demands a comprehensive understanding of diverse textual instructions and geometric concepts, the ability to locate the corresponding parts within the intricate structure of the CAD's SE operations, and the capability to generate concrete modifications to these SE operations.

\begin{figure*}
\begin{center}
\includegraphics[width=\textwidth]{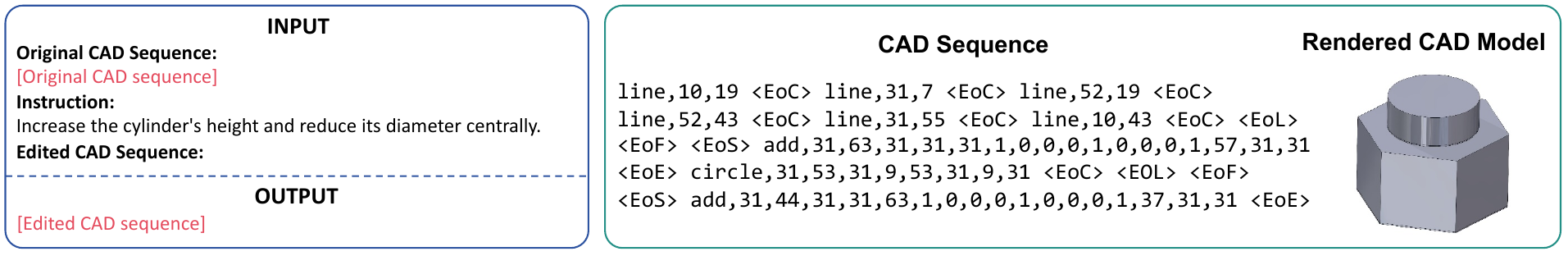}  
\end{center}
\caption{
\textbf{Left:} Example input and output for CAD-Editor. 
The input combines the original CAD sequence with the editing instruction, and the output is the edited CAD sequence. 
The specific CAD sequence is shortened to `[Original (or Edited) CAD Sequence]' to save space. 
\textbf{Right:} An illustration for a specific CAD sequence and its rendered CAD model.}
\label{fig:task_formulation}
\end{figure*}

\begin{figure*}[t]
\begin{center}
%\framebox[4.0in]{$\;$}
\includegraphics[width=\textwidth]{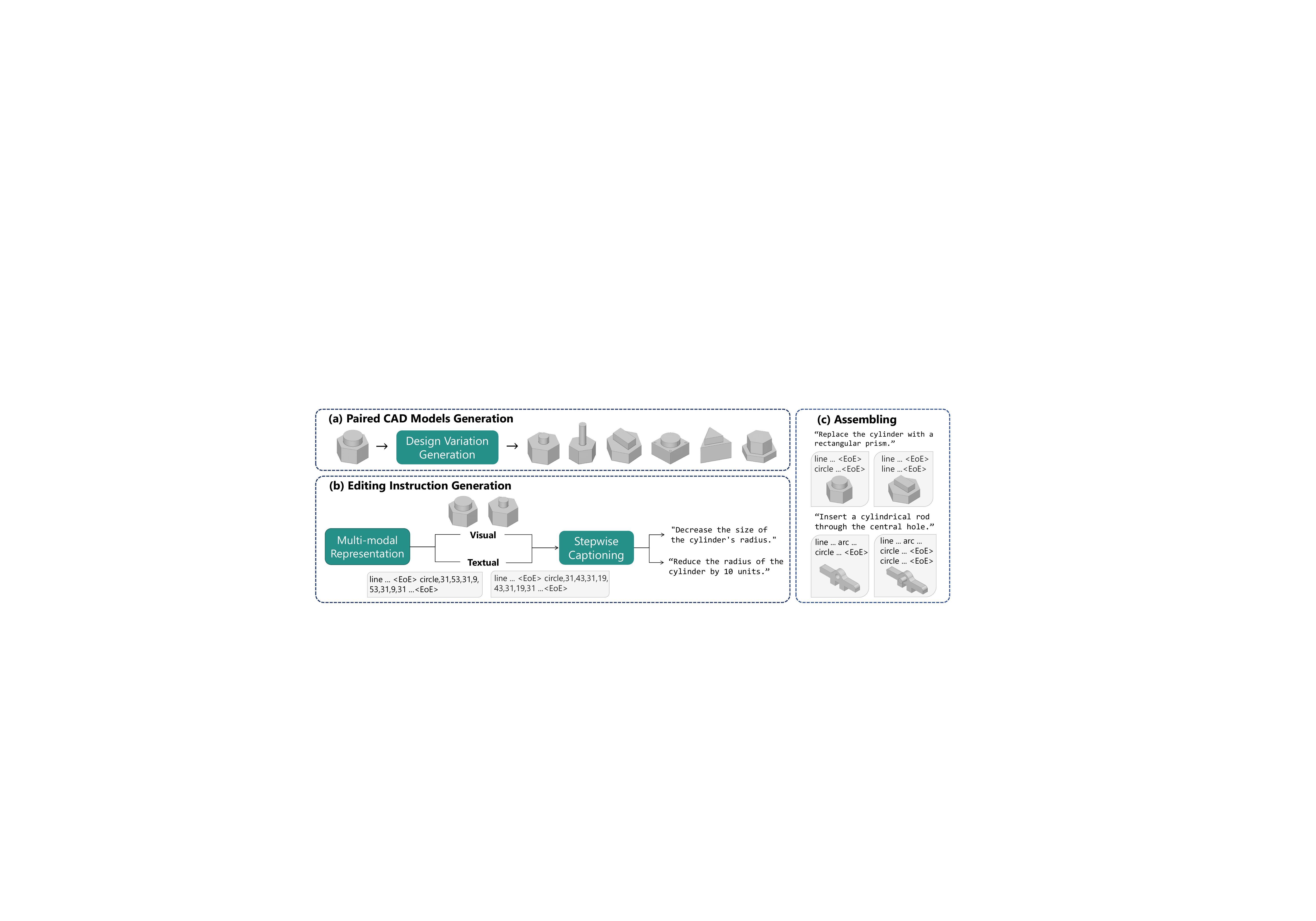}  
\end{center}
\caption{
Illustration of automated data synthesis pipeline.
}
\label{fig:data_generation}
\end{figure*}

In this work, we introduce \textbf{CAD-Editor}, the first framework for text-based CAD editing.
We frame the task as a sequence-to-sequence (seq2seq) generation problem, where the input combines an editing instruction and the sequence representation of the original CAD model, and the output is the sequence of the edited CAD model (Figure~\ref{fig:task_formulation}).
To address the need for triplet data with accurate correspondence, we propose an \emph{automated data synthesis pipeline} that leverages existing design variation models and Large Vision-Language Models (LVLMs) (Figure~\ref{fig:data_generation}).
Starting from an existing CAD model, design variation models generate edited CAD models by randomly altering parts while keeping others unchanged, producing pairs of original and edited CAD models.
LVLMs then summarize the differences between these CAD models into editing instructions, resulting in triplets with strong correspondence.
To tackle the composite nature of text-based CAD editing, we decompose the task into \emph{specialized sub-tasks: locating and infilling}, each addressing a specific aspect of the editing process (Figure~\ref{fig:framework}).
For both sub-tasks, Large Language Models (LLMs) serve as the backbone, leveraging their strong natural language understanding capabilities and basic CAD-related knowledge, including SE operations and geometric concepts~\citep{makatura2023can}.
In the locating stage, LLMs identify regions requiring modification by generating a masked CAD sequence, where special tokens \texttt{<mask>} indicate the regions to be modified.
In the infilling stage, LLMs generate appropriate edits for these masked regions, using the masked CAD sequence from the locating stage as context.

The contributions of this work are summarized as follows:
\begin{compactitem}  
    \item We introduce a new task, text-based CAD editing, enabling precise edits through textual instructions to better align with real-world user needs.
    \item We propose an automated data synthesis pipeline that combines design variation models and LVLMs to generate triplet data with accurate correspondence, addressing a critical challenge in training.
    \item We develop a locate-then-infill framework that decomposes the task into focused sub-tasks, and leverage LLMs as the backbone, effectively handling the composite nature of text-based CAD editing.
    \item We conduct extensive experiments, demonstrating that our approach outperforms baselines in generation validity, text-CAD alignment, and overall quality.
\end{compactitem}

\section{Related Works}\label{related_work}  
\noindent\textbf{CAD Generation.}  
Parametric CAD~\citep{shahin2008feature, camba2016parametric}, defined by its sketch-and-extrude operations, is central to mechanical design due to its ability to retain modeling history, which facilitates both editing and manufacturing. 
Recent large-scale CAD datasets~\citep{wu2021deepcad} have fueled the development of generative models.
\citet{wu2021deepcad} explored unconditional generation, where a random latent vector is used as input to generate CAD models. 
\citet{xu2022skexgen, xu2023hierarchical, zhang2024flexcad} focused on design variation generation, which randomly modifies specific part of an existing CAD model. 
Recently, \citet{khan2024cad, li2024cad} studied text-based CAD generation, which transforms textual descriptions into CAD models.
Our work differs in two key aspects.
First, we target a distinct task, unlike prior work, which either lacks text-based control~\citep{wu2021deepcad, xu2022skexgen, xu2023hierarchical, zhang2024flexcad} or disregards existing CAD models as constraints~\citep{khan2024cad, li2024cad}.
Second, we introduce a novel locate-then-infill framework based on LLMs to handle the composite nature of text-based CAD editing.
Previous approaches either rely on VAE-based~\citep{wu2021deepcad, xu2022skexgen, xu2023hierarchical} or transformer-based~\citep{khan2024cad, li2024cad} architectures, or apply LLMs without accounting for task-specific needs~\citep{zhang2024flexcad}.

\noindent\textbf{Large Language Models (LLMs).} 
Recently, LLMs like GPT~\citep{achiam2023gpt, OpenAI2023} and LLaMA~\citep{touvron2023llama} have distinguished themselves from smaller models, 
particularly through advanced prompting~\citep{brown2020language, wei2022chain, kojima2022large} or fine-tuning methods~\citep{weifinetuned}.
Beyond excelling in natural language processing~\citep{yue2024event, zhan2025coderagent, zhao2025evaluating}, LLMs have transform generative tasks in other domains, e.g. motion generation~\citep{zhang2024motiongpt} and material generation~\citep{gruverfine}. 
These advancements inspire us to adopt LLMs as the backbone for subtasks in our locate-then-infill framework.
Moreover, LLMs and LVLMs are increasingly utilized for data synthesis to enhance training~\citep{xu2024wizardlm, yumetamath}. 
Specifically, \citet{khantext2cad} leverage LLMs/LVLMs to synthesize data for text-based CAD generation. 
However, our task of text-based CAD editing presents distinct challenges.
First, unlike \citet{khan2024cad}, who generates two-tuple data (a text prompt and a CAD model), our task involves creating triplet data: an editing instruction, an original CAD model, and an edited CAD model. 
We address this by combining design variation models with LLMs/LVLMs.
Second, while \citet{khan2024cad} focus on captioning single CAD models, our task requires summarizing differences between two CAD models. 
We handle this by introducing a stepwise captioning strategy.

\noindent\textbf{Text-based Editing in Other Domains.}
Text-based editing has been widely explored across various domains, e.g., 3D editing~\citep{mikaeili2023sked}, image editing~\citep{meng2021sdedit, brooks2023instructpix2pix}, and video editing~\citep{chai2023stablevideo, ceylan2023pix2video}.
It enables users to specify and modify particular objects or attributes with precision and flexibility.
Inspired by these advancements, we introduce text-based editing in the CAD domain for the first time.

\section{Approach Overview}
\label{section:approach_overview}
% problem definition: I, C_orig, C_edit
Let $\mathcal{I}$ denotes the editing instruction, $\mathcal{C}_{\text{orig}}$ the original CAD model and $\mathcal{C}_{\text{edit}}$ the edited CAD model.
Here, the CAD model is represented using Sketch-and-Extrude (SE) operations, as this representation preserves the modeling history, making it easier to edit.
The goal of text-based CAD editing is to learn a function $f$ that takes $\mathcal{I}$ and $\mathcal{C}_{\text{orig}}$ as inputs and generates $\mathcal{C}_{\text{edit}}$ as output, i.e., $\mathcal{C}_{\text{edit}}=f(\mathcal{I},\mathcal{C}_{\text{orig}})$. 

% seq2seq proplem + sequence representation for C_orig, C_edit
We formulate text-based CAD editing as a sequence-to-sequence (seq2seq) generation problem. 
To achieve this, both the editing instruction and the CAD models are represented as sequences of textual tokens.
The editing instruction $\mathcal{I}$ naturally consists of textual tokens. 
For the CAD models $\mathcal{C}_{\text{orig}}$ and $\mathcal{C}_{\text{edit}}$, we adopt the sequence format introduced by \citet{zhang2024flexcad}, which abstracts all primitives in SE operations, including numerical and categorical parameters, into textual tokens (Figure~\ref{fig:task_formulation}).

To address the need for training data with good correspondence among $\mathcal{I}$, $\mathcal{C}_{\text{orig}}$, and $\mathcal{C}_{\text{edit}}$, denoted as $\mathbb{D} = \{(\mathcal{I}, \mathcal{C}_{\text{orig}}, \mathcal{C}_{\text{edit}})\}_{1}^{N}$ (where $N$ is the data size), we introduce an automated data synthesis pipeline (Section~\ref{section: dataset}).
First, we obtain $\mathcal{C}_{\text{edit}}$ from $\mathcal{C}_{\text{orig}}$ by leveraging existing design variation models~\citep{xu2023hierarchical} to randomly modify certain parts of the original CAD model while keeping others unchanged.
Next, we generate the corresponding $\mathcal{I}$ by utilizing LVLMs to summarize the differences between the paired CAD models.
Finally, we assemble these components into triplets.

Based on seq2seq formulation and triplet-correspondence dataset, we propose a locate-then-infill framework to tackle the composite nature of text-based CAD editing (Section~\ref{section: model}).
Specifically, we decompose the problem into two focused sub-tasks: locating and infilling, each handling a more manageable aspect of the overall problem. 
In the locating stage, we predict a masked CAD sequence $\mathcal{C}_{\text{mask}}$ by inserting special \texttt{<mask>} tokens into $\mathcal{C}_{\text{orig}}$, indicating the regions that require modification, i.e., $\mathcal{C}_{\text{mask}} = f_{\text{locate}}(\mathcal{I}, \mathcal{C}_{\text{orig}})$.
In the infilling stage, We generate $\mathcal{C}_{\text{edit}}$ by filling in precise modifications within the masked regions, i.e., $\mathcal{C}_{\text{edit}} = f_{\text{infill}}(\mathcal{I}, \mathcal{C}_{\text{orig}}, \mathcal{C}_{\text{mask}})$.
In both stages, LLMs serves as the backbone, leveraging their natural lauguage understanding and CAD knowledge acquired during the pre-training.

\section{Automated Data Synthesis Pipeline}
\label{section: dataset}
Figure~\ref{fig:data_generation} illustrates our data synthesis pipeline, comprising three key steps introduced below.

\noindent\textbf{Paired CAD Models Generation.}
In this step, we create paired CAD models, $\mathcal{C}_{\text{orig}}$ and $\mathcal{C}_{\text{edit}}$, by starting with an existing CAD model and applying design variation models to create its variations.
We source CAD models from the DeepCAD dataset~\citep{wu2021deepcad} and use Hnc-CAD~\citep{xu2023hierarchical} as our design variation model\footnote{We choose Hnc-CAD for its well-developed open-source implementation. Other design variation models~\citep{xu2022skexgen,zhang2024flexcad} are also applicable.}. 
Given an existing CAD model $\mathcal{C}_0$, we use Hnc-CAD’s auto-completion to generate variants $\mathcal{C}_1, \dots, \mathcal{C}_K$.
We then create paired CAD models by: 1) treating $\mathcal{C}_0$ as the original and $\mathcal{C}_k$ ($k \in [1, K]$) as the edited CAD model; 2) reversing the roles, considering $\mathcal{C}_k$ as the original and $\mathcal{C}_0$ as the edited CAD model; and 3) using two generated variants, $\mathcal{C}_{k_1}$ and $\mathcal{C}_{k_2}$ ($k_1, k_2 \in [1, K], k_1 \neq k_2$), as the original and edited CAD models.
This approach captures common CAD editing operations, including addition, deletion, and replacement.

\noindent\textbf{Editing Instruction Generation.}
In this step, we generate editing instructions, $\mathcal{I}$, by summarizing the difference between $\mathcal{C}_{\text{orig}}$ and $\mathcal{C}_{\text{edit}}$ using LVLMs.
To ensure both diversity and accuracy, we introduce two key techniques.

First, we represent CAD models in \emph{multiple modalities}: the visual and sequence modalities. 
The intuition is that high-level structural changes (e.g., changing a cylinder to a cube) are more easily observed in the visual modality, while low-level numerical changes (e.g., doubling the height) are better captured in the sequence modality. Additionally, the sequence modality provides a detailed representation of all operations, ensuring that no information is lost.
In our implementation, we use rendered images for the visual modality and SE operations for the sequence modality. 

Second, we propose a \emph{stepwise captioning strategy} to break down the complex task of difference summarization into three sub-tasks, enhancing generation quality of LVLMs.
For each representation (visual or sequence) of a CAD model pair, we follow these steps: 
1) describing each CAD model – this involves analyzing geometric properties such as component types, quantities, sizes, and spatial relationships;
2) identifying differences – using the detailed descriptions from the previous step alongside CAD models, this stage extracts the necessary modifications between CAD models;
and 3) compressing instructions – the final step refines the editing instructions into a concise yet precise form.

\noindent\textbf{Assembling.}
Finally, we assemble CAD pairs ($\mathcal{C}_{\text{orig}}$ and $\mathcal{C}_{\text{edit}}$) from the first step and editing instruction ($\mathcal{I}$) from the second step into triplets, constructing the training dataset $\mathbb{D} = \{(\mathcal{I}, \mathcal{C}_{\text{orig}}, \mathcal{C}_{\text{edit}})\}_{1}^{N}$.
Note that the visual modality is only used in the second stage to generate diverse editing instructions.
In the final dataset, all CAD models are represented as SE operations, aligning with the focus of this work – text-based CAD editing in the SE domain.

\begin{figure*}[t]
\begin{center}
\includegraphics[width=\textwidth]{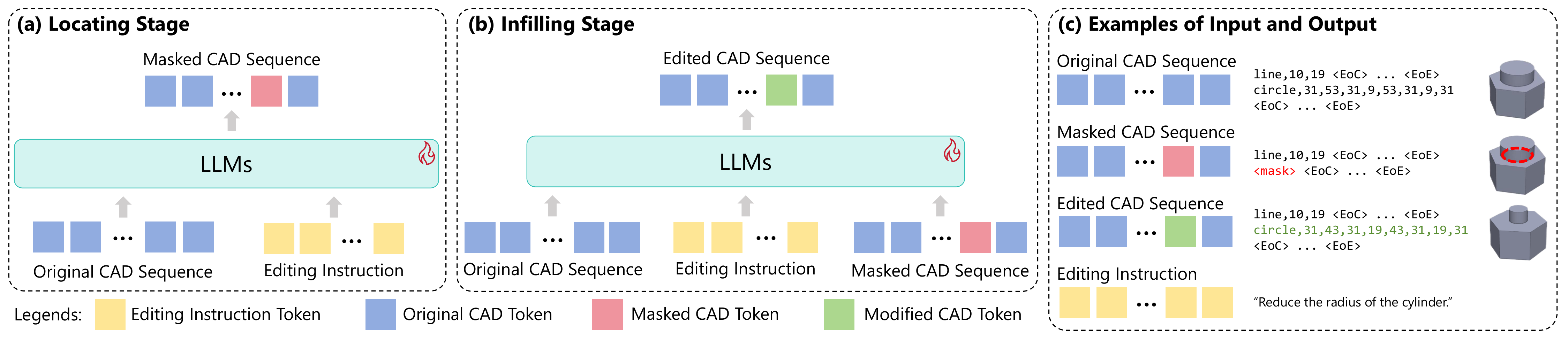}  
\end{center}
\caption{
\textbf{(a)-(b)}: Overview of Locate-then-Infill framework.
\textbf{(c)}: Examples of input and output, where the left column shows abstracted representations using legends, the middle column displays concrete sequences and the right column presents rendered visual objects.
}
\label{fig:framework}
\end{figure*}

\section{Locate-then-Infill Framework}
\label{section: model}
Figure~\ref{fig:framework} illustrates our framework. 
We decompose text-based CAD editing by explicitly introducing a masked CAD sequence, $\mathcal{C}_{\text{mask}}$, to indicate potential modification regions:
\begin{align}
         & P(\mathcal{C}_{\text{edit}}\mid \mathcal{C}_{\text{orig}},\mathcal{I})
          \triangleq \\\nonumber
        & P(\mathcal{C}_{\text{mask}}\mid \mathcal{C}_{\text{orig}},\mathcal{I})\cdot P(\mathcal{C}_{\text{edit}} \mid \mathcal{C}_{\text{orig}},\mathcal{I},\mathcal{C}_{\text{mask}}),
\end{align}
where $P(\cdot)$ denotes the probability.
Here, $P(\mathcal{C}_{\text{mask}}\mid\mathcal{C}_{\text{orig}},\mathcal{I})$ represents locating stage, 
while $P(\mathcal{C}_{\text{edit}}\mid\mathcal{C}_{\text{orig}},\mathcal{I},\mathcal{C}_{\text{mask}})$ corresponds to the infilling stage.

\subsection{Locating Stage}
This stage aims to generate a masked CAD sequence, $\mathcal{C}_{\text{mask}}$, where regions requiring modification are marked by special tokens \texttt{<mask>} while unchanged parts are copied from the original CAD sequence $\mathcal{C}_{\text{orig}}$.
We adopt LLMs as the backbone and autoregressively predict tokens in $\mathcal{C}_{\text{mask}}$ using $\mathcal{C}_{\text{orig}}$ and the editing instruction $\mathcal{I}$ as context:
\begin{equation}
P(\mathcal{C}_{\text{mask}} \mid \mathcal{C}_{\text{orig}}, \mathcal{I}) = \prod_{t=1}^T P(c^t_{\text{mask}} \mid \mathcal{C}_{\text{orig}}, \mathcal{I}, c^{<t}_{\text{mask}}),
\end{equation}
where $T$ is the sequence length.

\noindent\textbf{Creating Ground-Truth Masked CAD Sequence via LCS.}
To finetune LLMs for the locating task, we require ground-truth masked CAD sequences, denoted as $\mathcal{C}_{\text{gt-mask}}$ as supervision signals.
We obtain them using the \emph{Longest Common Subsequence} (LCS) algorithm.
Let \( \mathcal{C}_{\text{orig}} = \{c^1,\dots,c^T\} \) and \( \mathcal{C}_{\text{edit}} = \{\tilde{c}^1, \dots, \tilde{c}^T\} \) denote the tokenized original and edited CAD sequences respectively.
The LCS algorithm computes the longest subsequence \( \mathcal{C}_{\text{LCS}} = \{c^{i_1}, \dots, c^{i_k}\} \) such that \( \mathcal{C}_{\text{LCS}} \subseteq \mathcal{C}_{\text{orig}} \) and \( \mathcal{C}_{\text{LCS}} \subseteq \mathcal{C}_{\text{edit}} \), where the indices \( i_1, i_2, \dots, i_k \) represent the positions of matching tokens in \( \mathcal{C}_{\text{orig}} \).
Using \( \mathcal{C}_{\text{LCS}} \), we construct $\mathcal{C}_{\text{gt-mask}}$ as follows: 1) for each token \( c^i \in \mathcal{C}_{\text{orig}} \),
if \( c^i \in \mathcal{C}_{\text{LCS}} \), retain \( c^i \) in \( \mathcal{C}_{\text{gt-mask}} \) and if \( c^i \notin \mathcal{C}_{\text{LCS}} \), replace \( c^i \) with the placeholder token \texttt{<mask>}; 2) for tokens in \( \mathcal{C}_{\text{edit}} \) that do not appear in \( \mathcal{C}_{\text{LCS}} \) (representing insertions), insert \texttt{<mask>} tokens at the corresponding positions in \( \mathcal{C}_{\text{gt-mask}} \); 3) consecutive \texttt{<mask>} tokens are merged into a single \texttt{<mask>} token for simplicity.

\subsection{Infilling Stage}
\label{subsec:infill}
This stage focuses on generating the final edited sequence \( \mathcal{C}_{\text{edit}} \) by precisely filling in the masked regions while preserving the unmodified parts. 
We employ LLMs as the backbone, autoregressively predicting tokens in \( \mathcal{C}_{\text{edit}} \) using \( \mathcal{C}_{\text{mask}} \) from the locating stage along with $\mathcal{C}_{\text{orig}}$ and $\mathcal{I}$ as inputs:
\begin{align}
& P(\mathcal{C}_{\text{edit}} \mid \mathcal{C}_{\text{mask}}, \mathcal{C}_{\text{orig}}, \mathcal{I}) \\\nonumber
 = & \prod_{t=1}^T P(c^t \mid \mathcal{C}_{\text{mask}}, \mathcal{C}_{\text{orig}}, \mathcal{I}, c^{<t}).
\end{align}

\noindent \textbf{Improving Performance with Selective Data.}
To fine-tune LLMs for the infilling task, we use the dataset $\mathbb{D}$ synthesized through the pipeline introduced in Section~\ref{section: dataset}. 
While we strive to ensure high-quality synthetic data, achieving absolute accuracy is impossible.
To further improve performance, we introduce a \emph{selective dataset} curated with human annotations. 
Rather than having human annotators create editing instructions or CAD models from scratch, we adopt a more efficient approach — inviting them to select the best option from generated candidates.
This significantly reduces human effort and accelerates dataset construction.
Specifically, we first fine-tune LLMs on synthetic data and use them to generate multiple edited CAD sequences. 
These sequences are then rendered into visual objects, and human annotators select the best one. 
The chosen sequence, along with its corresponding original CAD model and textual instruction, is added to the selective dataset.
Finally, we further fine-tune LLMs using this selective dataset.

\subsection{Training and Inference}
\noindent\textbf{Training.}
In the locating stage, we fine-tune LLMs using Low-Rank Adapters (LoRA) \citep{hulora} with the ground-truth masked CAD sequence constructed via LCS.
For the infilling stage, we first fine-tune LLMs using LoRA with the synthetic dataset introduced in Section~\ref{section: dataset}. 
We then further refine the model by fine-tuning it with LoRA on the selective dataset introduced in Section~\ref{subsec:infill}.

\noindent\textbf{Inference.}
During inference, the locating and the infilling stage operates sequentially.
The locating stage generates \( \mathcal{C}_{\text{mask}} \) using \( \mathcal{C}_{\text{orig}} \) and \( \mathcal{I} \) as input.
The infilling stage generates \( \mathcal{C}_{\text{edit}} \) by using \( \mathcal{C}_{\text{mask}} \) from the locating stage along with \(\mathcal{C}_{\text{orig}} \) and \( \mathcal{I} \) as input.

\section{Experiments}
\label{experiments}

\begin{figure*}[t]
\begin{center}
\includegraphics[width=\textwidth]{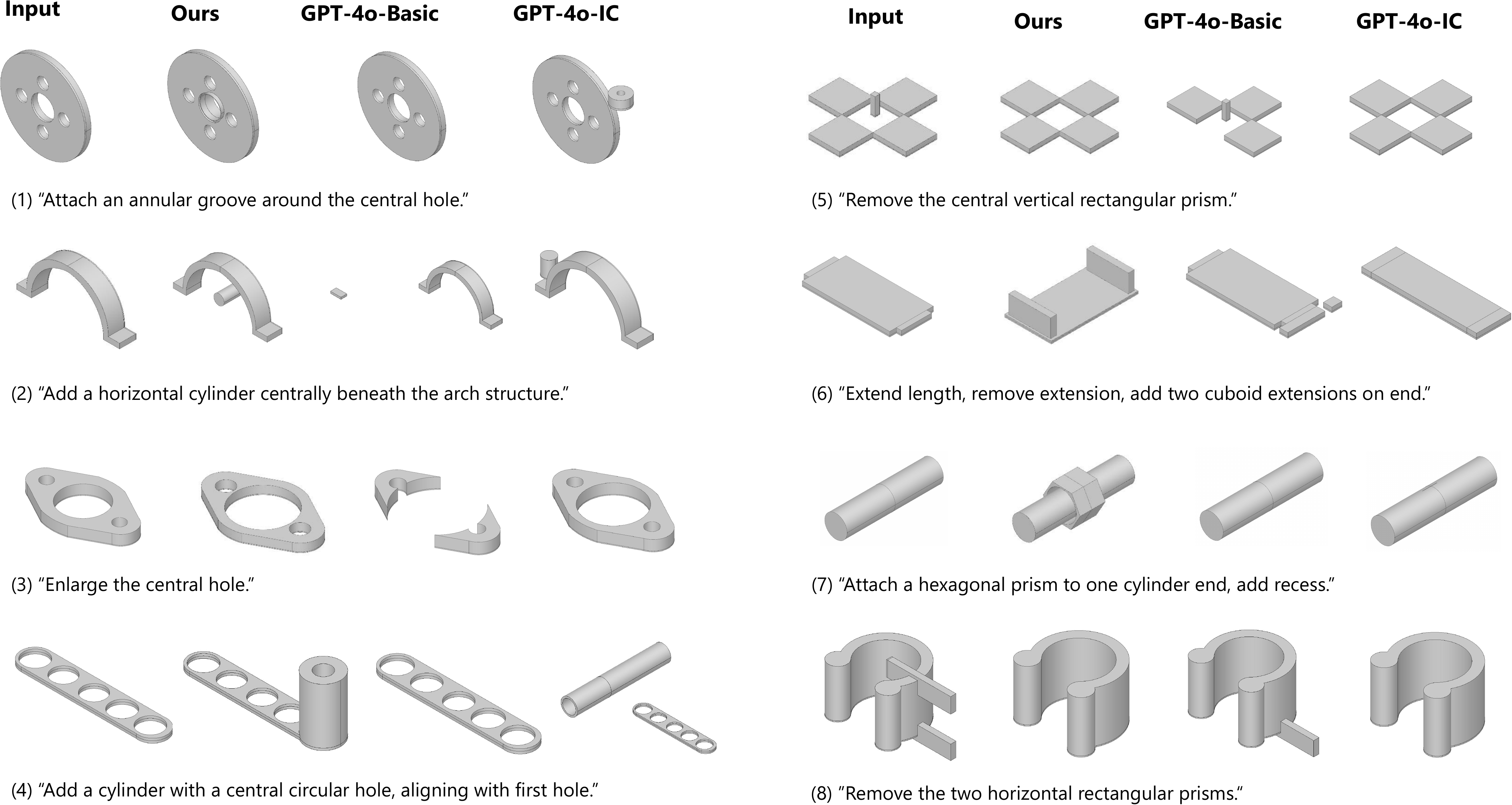}  
\end{center}
\caption{Qualitative results from CAD-Editor, GPT-4o-Basic and GPT-4o-IC .}
\label{comparison}
\end{figure*}
  
\subsection{Experimental Setup}

\noindent\textbf{Datasets.}
We use the DeepCAD dataset~\citep{wu2021deepcad} which contains 178k CAD models.
We split it into 90\% training, 5\% validation, and 5\% testing segments. 
We follow the same strategy in existing work~\citep{xu2022skexgen,xu2023hierarchical} to remove duplication.
For the synthetic dataset used for training, we generate 120k examples using the method introduced in Section~\ref{section: dataset}. 
For the test set, we randomly sample 2k examples from the original test segment, generate the initial version following Section~\ref{section: dataset}, and manually examine them to ensure the correctness.
To compare the performance of different methods, we generate 5 outputs for each example in the test set, yielding 10k CAD models for evaluation.

\noindent\textbf{Implementation Details.}
\label{implementation}
During data synthesis, GPT-4o is utilized for the visual modality, while LLaMA-3-70B handles the sequence modality. For training, we adopt Llama-3-8B-Instruct as the backbone model, fine-tuning it over 70 epochs using PyTorch's Distributed Data Parallel (DDP) framework on 4 NVIDIA A800-80GB SMX GPUs.
The initial learning rate is set to 1e-4 with a maximum token length of 1024. 
We employ LoRA with a rank of 32. 
During the inference, we set the temperature as 0.9 and top-p as 0.9 to generate varied results in each trial.

\noindent\textbf{Baselines.}
We compare our results with three types of baselines: 1) design variation generation models, including \textbf{SkexGen}~\citep{xu2022skexgen}, \textbf{Hnc-CAD}~\citep{xu2023hierarchical} and \textbf{FlexCAD}~\citep{zhang2024flexcad}; 2) text-based CAD generation models such as \textbf{Text2CAD}~\citep{khan2024cad} and 3) foundation models that are not specifically designed for CAD generation but have acquired some CAD knowledge during pre-training.
For the third category, we select one of the most powerful foundation models, GPT-4o, as our baseline. 
We use two prompting strategies: 1) \textbf{GPT-4o-Basic}, which provides only an explanation of CAD operation sequences; and 2) \textbf{GPT-4o-IC}: which includes three in-context (IC) examples retrieved based on cosine similarity between editing instructions, in addition to the basic explanation.
Notably, existing CAD design variation models and text-based CAD generation models do not support text-based CAD editing. 
We include them as baselines to show that our model can generate CAD models with comparable or superior validity and quality while addressing a more complex task.

\begin{figure*}[t]
\begin{center}
\includegraphics[width=\textwidth]{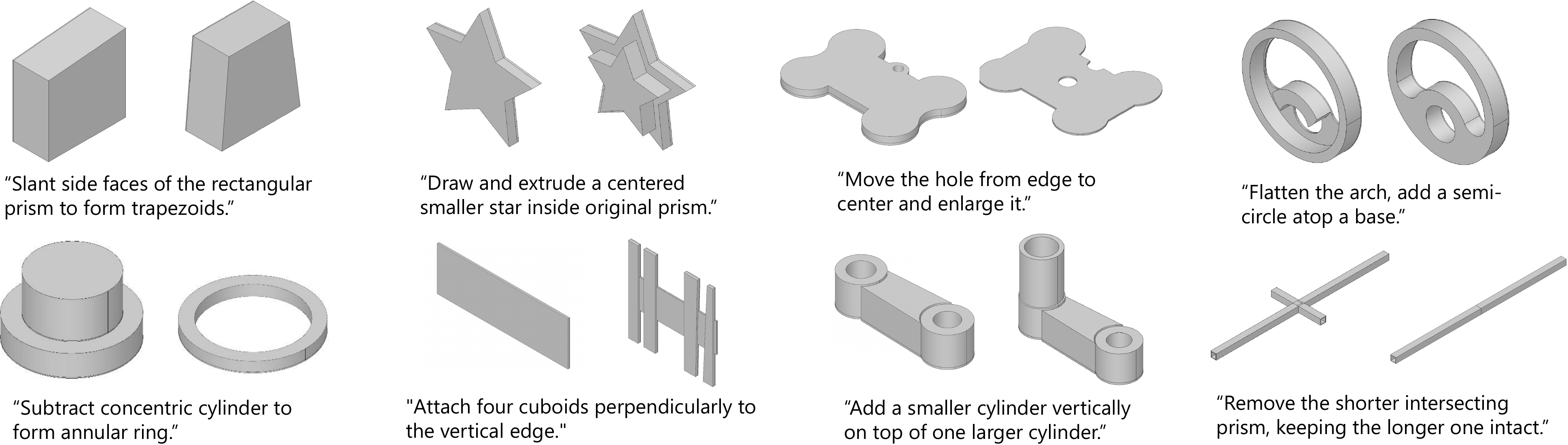}  
\end{center}
\caption{
Additional results from CAD-Editor with various editing instructions .
In each sub-figure, the left image shows the original CAD model, the right image displays the edited CAD model and the text below provides the editing instruction.
}
\label{diverse}
\end{figure*}

\begin{figure*}[t]
\begin{center}
\includegraphics[width=0.95\textwidth]{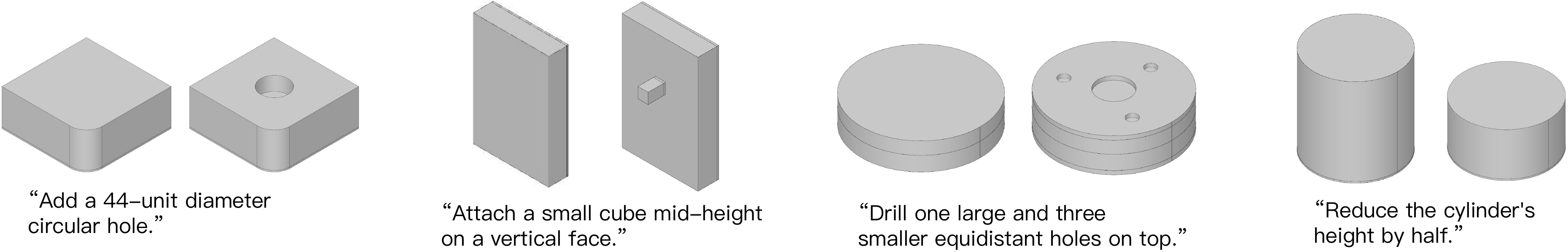}  
\end{center}
\caption{CAD-Editor can deal with parametric instructions, including explicit numeric expressions and implicit parametric cues.}
\label{fig:parameter}
\end{figure*}

\subsection{Metrics}
As this work is the first to address text-based CAD editing, we propose evaluating the task from three key aspects.

\noindent \textbf{(1) Validity. } The generated CAD sequence must be valid, that is, it can be successfully parsed and rendered into a 3D visual object. 
We denote this as \textbf{Valid Ratio (VR)}.

\noindent \textbf{(2) Realism}. The generated CAD models should be realistic and similar to ground-truth CAD models. 
To measure this, we adopt the \textbf{Jensen-Shannon Divergence (JSD)} from prior work~\citep{wu2021deepcad, xu2022skexgen, xu2023hierarchical}. 
JSD quantifies the similarity between two probability distributions, evaluating how often the ground-truth point clouds occupy similar positions as the generated point clouds.

\textbf{(3) Edit Consistency}. 
The generated edits should faithfully reflect the provided textual instructions. We assess consistency at both the point cloud and image levels:

\noindent $\bullet$ \textit{Point Cloud Level.}
We use the \textbf{Chamfer Distance (CD)} as a reconstruction metric as used in \cite{khantext2cad} to evaluate geometric alignment between the edited and ground-truth CAD model pairs. While CD captures shape similarity, it may not fully reflect semantic consistency, as the goal of editing is not exact replication but meaningful transformation aligned with the instruction.

\noindent $\bullet$ \textit{Image Level.}
We adapt  \textbf{Directional CLIP Score (D-CLIP)} from the image editing domain~\citep{gal2022stylegan, brooks2023instructpix2pix} as the metric.
Building upon CLIP score~\citep{radford2021learning, sohn2023styledrop}, D-CLIP evaluates how well the change between the image for the edited CAD model and the image for the original CAD model aligns with the editing instruction:
\begin{align}
    & \quad\quad\quad\quad\quad\quad
    \text{D-CLIP} = \frac{\Delta I \cdot \Delta T}{\lvert \Delta I \rvert \lvert \Delta T \rvert},\\\nonumber\\
    & \Delta T  = E_T\left(t_{\text{edit}}\right) - E_T\left(t_{\text{orig}}\right), 
    \Delta I = E_I\left(i_{\text{edit}}\right) - E_I\left(i_{\text{orig}}\right),\nonumber
\end{align}
where $E_I$ and $E_T$ are CLIP's image and text encoders, respectively. $t_{\text {orig}}$ is a neutral text (e.g.,``This is a 3D shape.''),  $t_{\text {edit}} $ is the concatenation of $t_{\text {orig}}$ and the textual instruction. $i_{\text {edit }}$ and $i_{\text {orig }}$ are the images for the edited and original CAD model, respectively.

\begin{table}[t]  
\caption{
Quantitative evaluations. SkexGen, Hnc-CAD, FlexCAD and Text2CAD do not support text-based editing, so only their generation quality is compared. JSD, CD, and D-CLIP values are scaled by $10^2$. $\uparrow$: the higher the better,  $\downarrow$: the lower the better.
}  
\label{tab:baselines}
\begin{small}
\begin{center}  
\resizebox{\linewidth}{!}{
\begin{tabular}{lcccccc}  
\toprule  
\multicolumn{1}{l}{\bf Method}  & \multicolumn{1}{c}{\bf VR $\uparrow$ }  &  \multicolumn{1}{c}{\bf JSD $\downarrow$} &  \multicolumn{1}{c}{\bf  CD $\downarrow$} &  \multicolumn{1}{c}{\bf D-CLIP $\uparrow$}  & \multicolumn{1}{c}{\bf H-Eval $\uparrow$} \\  
\midrule  
\textbf{SkexGen}       & 74.3  & 1.94 & - & -  & - \\  
\textbf{Hnc-CAD}      & 77.4  & 1.77  & -& -  & - \\ 
\textbf{FlexCAD}      & 82.1 &  1.72  & - &-  & - \\ 
\textbf{Text2CAD}      & 84.8   & 2.39 & 1.91 &-   & - \\ 
\midrule
\textbf{GPT-4o-Basic}  & 63.2  & 1.10  & 2.30  &- 1.08 & 7.22 \\  
\textbf{GPT-4o-IC}  & 84.5 & 0.70  & 1.55  &- 0.11  & 15.6 \\  
\textbf{CAD-Editor}   & \textbf{95.6}  &  \textbf{0.65}   & \textbf{1.18}  & \textbf{0.11} & \textbf{43.2} \\    
\bottomrule  
\end{tabular}
}
\end{center}
\end{small}
\end{table}

\subsection{Main Results}

\noindent\textbf{Quantitative Evaluation.}
Table \ref{tab:baselines} reports the average scores across 3 runs. 
Notably, CAD-Editor achieves a high Valid Ratio of 95.6\%, significantly surpassing other methods and indicating a greater proportion of valid and high-quality CAD generations. 
In terms of CD and D-CLIP, which measure alignment with editing instructions, CAD-Editor achieves scores of 1.18 and 0.11 respectively, representing substantial improvements over both GPT-4o-Basic and GPT-4o-IC. These results underscore CAD-Editor's effectiveness in adhering to user instructions. 
Additionally, CAD-Editor outperforms all baselines on the point cloud evaluation metric JSD, demonstrating good generation quality.  
Overall, the results indicate that CAD-Editor not only enable precise text-based CAD editing, which better alignment with user instructions, but also achieve higher validity and better quality of the generated CAD designs.

\noindent\textbf{Qualitative Evaluation.}
In Figure \ref{comparison}, we qualitatively compare our method with GPT-4o-Basic and GPT-4o-IC. We observe that GPT-4o-Basic often generates irrelevant edits (case 6), unrealistic shapes (case 3), or fails to make any changes (cases 1 ,4 and 7). Additionally, it struggles with distinguishing shape types (case 2) and locating specific positions (case 5).  It performs better with dynamic few-shot prompting (i.e., GPT-4o-IC), highlighting the high quality of our synthetic data. GPT-4o-IC can detect specific shapes reasonably well but still struggles with precise localization (case 2 and 4) and object count (case 8). In contrast, our model successfully executes many challenging edits, including modifying sizes, shapes, and positions, as well as replacing, adding, and removing objects.

\begin{figure}[t]
\begin{center}
\includegraphics[width=0.48\textwidth]{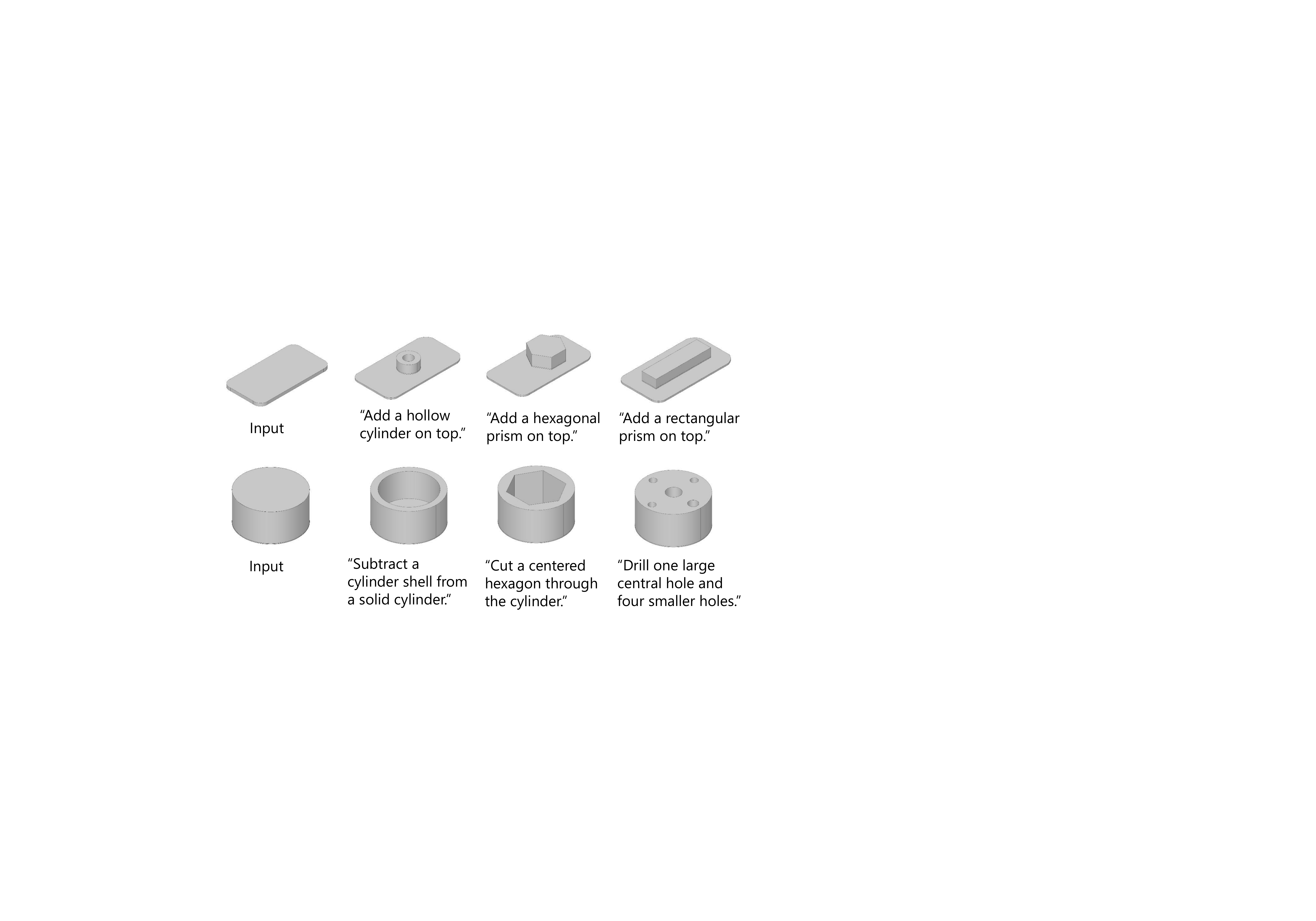} 
\end{center}
\caption{Given one CAD model and various instructions, CAD-Editor produces different outcomes.}
\label{multiple}
\end{figure}

\begin{figure}[t]
\begin{center}
\includegraphics[width=0.48\textwidth]{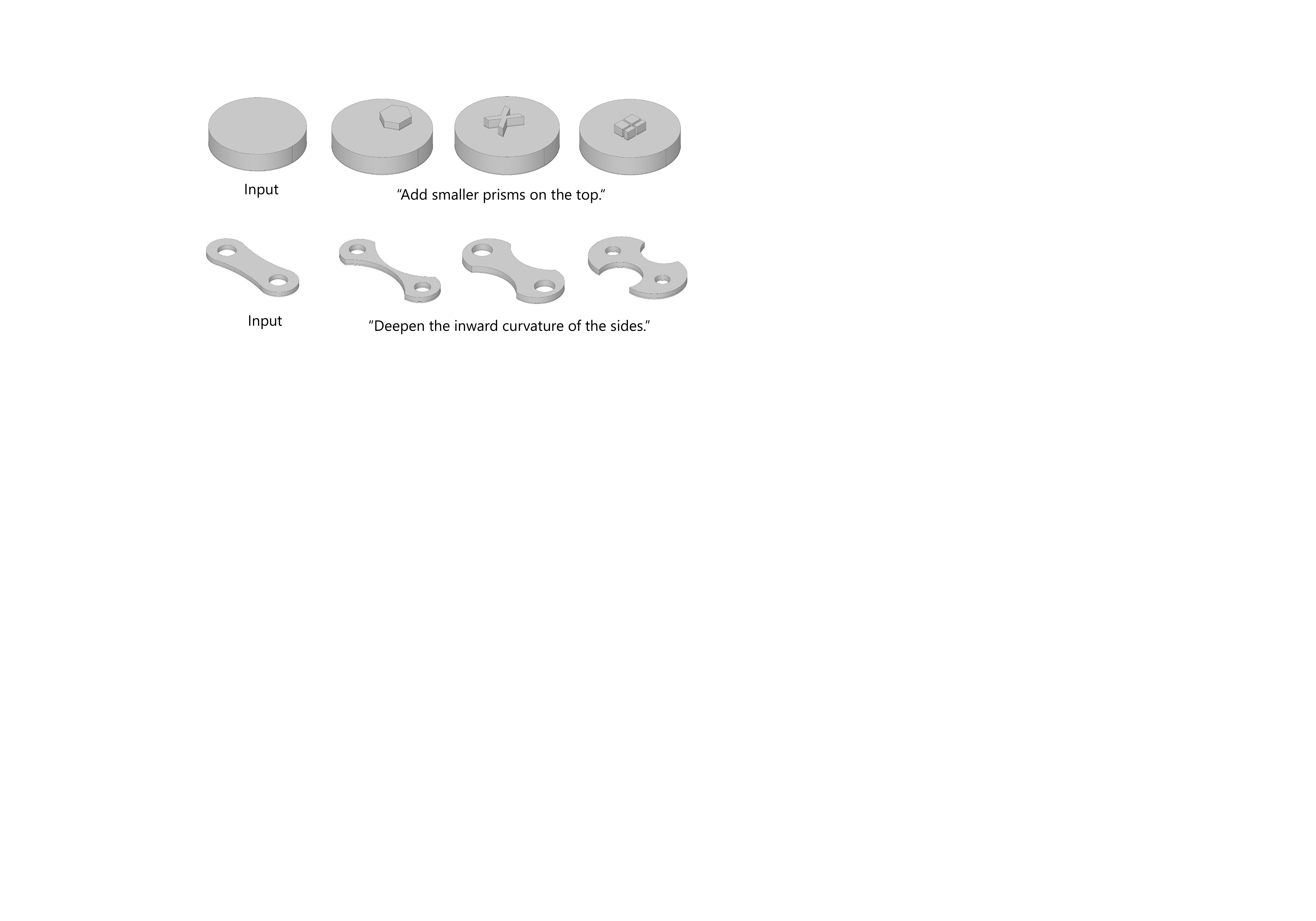} 
\end{center}
\caption{Given the same CAD model and instruction, CAD-Editor produces diverse outcomes.}
\label{onetomore}
\end{figure}

\begin{figure*}[t]
\begin{center}
\includegraphics[width=\textwidth]{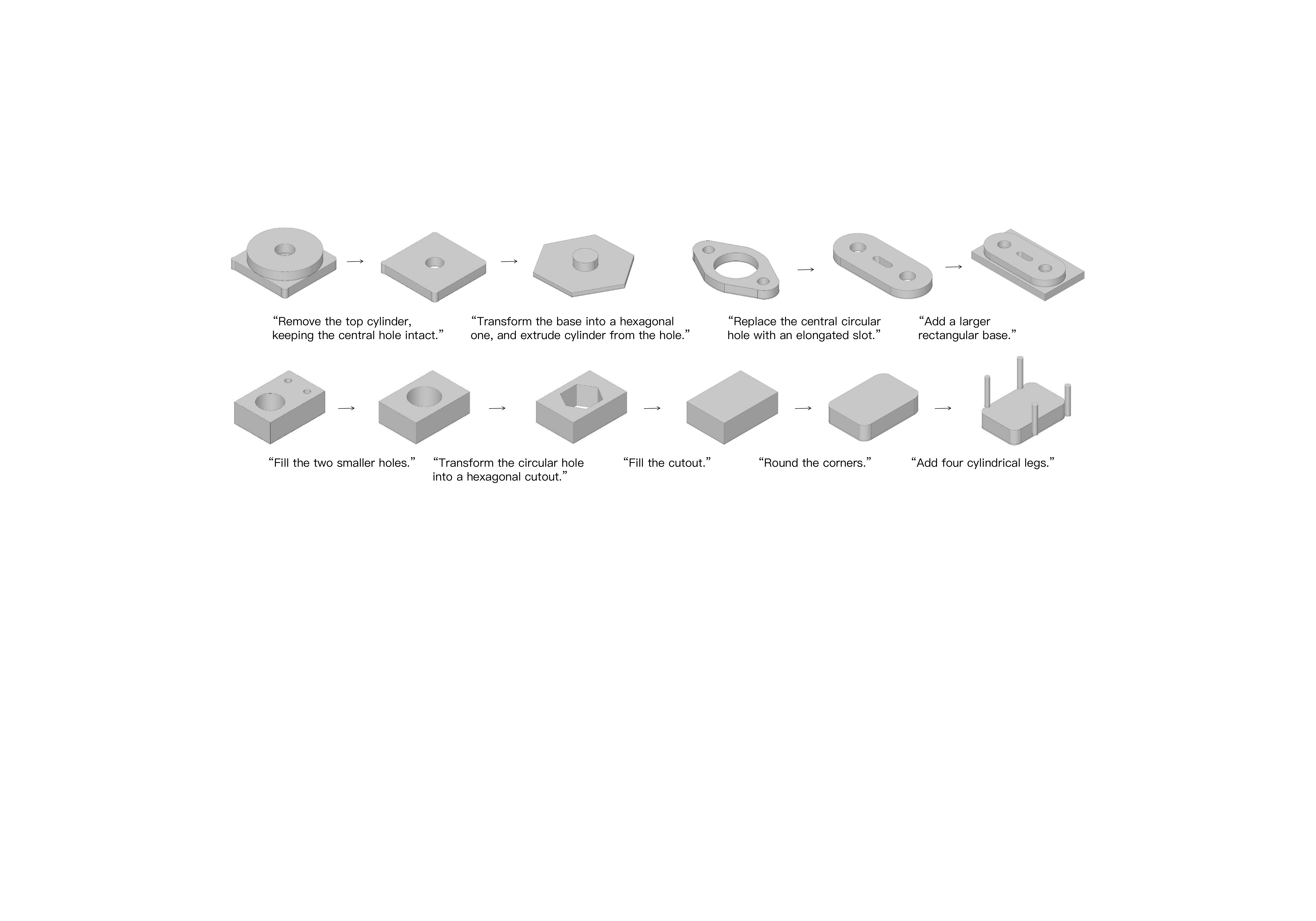}  
\end{center}
\caption{Apply CAD-Editor iteratively to edit a CAD model until it meets user requirements. }
\label{continuous}
\end{figure*}

Besides, we present more results from CAD-Editor.
Figure~\ref{diverse} shows how CAD-Editor handles a variety of editing instructions and different CAD models.
Figure~\ref{fig:parameter} shows that CAD-Editor has the ability to interpret parameterized instructions, including explicit numeric expressions such as ``add a 44-unit diameter circular hole'' and implicit parametric cues like ``reduce the cylinder's height by half''.
Figure~\ref{multiple} illustrates that, given the same original CAD model, CAD-Editor applies different modifications based on the provided editing instructions.
Figure~\ref{onetomore} shows that when the editing instruction is vague, CAD-Editor generates multiple CAD models, all aligning with the user’s intent.
Figure~\ref{continuous} highlights the iterative editing capability of CAD-Editor, allowing users to refine a CAD model through successive instructions until a satisfactory result is achieved.

\noindent\textbf{Human Evaluation.}
We randomly sampled 2,000 CAD models from the full set of generated results. 
Each pair of original CAD model and edited CAD model was independently rated by five crowd workers. 
For each pair, a score of 1 is assigned if the generated data is deemed successful, and 0 otherwise. 
Success is defined by two criteria: alignment with the text and sufficiently high visual quality. 
The results, denoted as \textbf{H-Eval}, are presented in Table \ref{tab:baselines}.
CAD-Editor outperforms baselines, indicating that crowd workers frequently found CAD models generated by baselines to be misaligned with the instructions or of lower quality, whereas our method demonstrated superior performance.

\begin{table}[t]  
\caption{Ablation studies. The CAD-Editor-mini is trained on a small set with 10k examples.}  
\label{ablation}
\begin{small}
\begin{center}  
\resizebox{\linewidth}{!}{
\begin{tabular}{lcccc}  
\toprule  
\multicolumn{1}{c}{\bf Method} & \multicolumn{1}{c}{\bf VR $\uparrow$}  &   \multicolumn{1}{c}{\bf JSD $\downarrow$}&  \multicolumn{1}{c}{\bf  CD $\downarrow$} & 
 \multicolumn{1}{c}{\bf D-CLIP $\uparrow$}  \\  
\midrule
CAD-Editor-mini w/ Basic & 88.8  &  0.78 & 1.25 & - 0.12 \\
CAD-Editor-mini w/ Step & 90.1 & 0.70  & 1.22 & - 0.07 \\
\midrule  
CAD-Editor w/ Direct  & 86.1 &  0.77 & 1.26 & - 0.19  \\  
CAD-Editor w/ L-I  & 96.5  &  0.67 & 1.13 & 0.03  \\ 
CAD-Editor w/ L-I \& HS &  95.6  & 0.65 & 1.18  & 0.11  \\ 

\bottomrule  
\end{tabular}  
}
\end{center}  
\end{small}
\end{table} 

\subsection{Ablation Studies}
\noindent\textbf{Stepwise Captioning Strategy.}
As introduced in Section~\ref{section: dataset}, we propose a stepwise captioning strategy to decompose the complex task of generating editing instructions.
To evaluate its effectiveness, we conduct an experiment where LVLMs are directly queried to generate editing instructions, denoted as \textbf{CAD-Editor w/ Basic}.
Our approach, which incorporates stepwise captioning, is referred to as \textbf{CAD-Editor w/ Step}.
Due to resource constraints, we compare these methods on a subset of 10k examples.
Table~\ref{ablation} shows results.
CAD-Editor w/ Step outperforms CAD-Editor w/ Basic across all metrics, highlighting the importance of stepwise captioning strategy in ensuring accurate  editing instruction generation.

\noindent\textbf{Locate-then-Infill Framework.}
Our approach consists of two stages: the locating and infilling stages, each focused on a specialized sub-task of text-based CAD editing.
A more straightforward approach would be to treat the task as a whole, without decomposing it, and directly fine-tune LLMs.
We refer to this as \textbf{CAD-Editor w/ Direct} and compare it with our method, \textbf{CAD-Editor w/L-I}.
As shown in Table~\ref{ablation}, CAD-Editor w/L-I outperforms CAD-Editor w/ Direct, showing the effectiveness of our two-stage approach in addressing the composite nature of text-based CAD editing.

\noindent\textbf{Selective Dataset.}
In Section~\ref{subsec:infill}, we propose improving the performance of the infilling stage with selective data curated with human annotation.
We conduct experiments both with or without such human selection, denoted as \textbf{CAD-Editor w/ L-I \& HS} and \textbf{CAD-Editor w/ L-I}.
Table \ref{ablation} shows that this strategy results in the greatest improvement in both JSD and D-CLIP scores, demonstrating its effectiveness in enhancing generation quality and better aligning editing instructions with the edited CAD model.

\section{Limitation}
While CAD-Editor shows promising results, it has limitations.
First, its data synthesis pipeline relies on LVLMs, which are costly and struggle with processing multiple images simultaneously.
Second, as an LLM-based system, it faces challenges with long contexts and generating extended sequences, limiting its ability to handle highly complex CAD models and their corresponding edits.

\section{Conclusion}
We introduced CAD-Editor, the first framework for text-based CAD editing.
We proposed a data synthesis pipeline to address the need of triplet data with accurate correspondence, and a locate-then-infill framework to handle the composite nature of the task.
Experiments showed that CAD-Editor outperforms other methods.
In the future, we aim to enhance our data synthesis pipeline to make it more cost-effective and efficient. 
We also plan to develop an advanced benchmark that better reflects practical user scenarios.

\section*{Acknowledgements}

We thank Ruiyu Wang for his assistance in conducting one of the baseline experiments.
We also appreciate the constructive comments from the anonymous reviewers.

\section*{Impact Statement}  
This work presents a novel task: text-based CAD editing, along with an automated data annotation pipeline and a locate-then-infill framework to effectively address it. By streamlining the CAD model development process, our approach has the potential to significantly accelerate design workflows and make CAD modeling more accessible to a broader audience, especially those with limited design expertise. Rather than replacing human designers, our goal is to augment creativity and productivity, empowering individuals to bring their ideas to life more efficiently and effectively.

\section*{Ethics Statement}
The data used in this work is tailored for creating and modifying CAD models. Due to its specialized nature, the misuse risk is naturally minimized, ensuring that the developed methods primarily benefit design and engineering tasks.

\bibliography{example_paper}
\bibliographystyle{icml2025}

%%%%%%%%%%%%%%%%%%%%%%%%%%%%%%%%%%%%%%%%%%%%%%%%%%%%%%%%%%%%%%%%%%%%%%%%%%%%%%%
%%%%%%%%%%%%%%%%%%%%%%%%%%%%%%%%%%%%%%%%%%%%%%%%%%%%%%%%%%%%%%%%%%%%%%%%%%%%%%%
% APPENDIX
%%%%%%%%%%%%%%%%%%%%%%%%%%%%%%%%%%%%%%%%%%%%%%%%%%%%%%%%%%%%%%%%%%%%%%%%%%%%%%%
%%%%%%%%%%%%%%%%%%%%%%%%%%%%%%%%%%%%%%%%%%%%%%%%%%%%%%%%%%%%%%%%%%%%%%%%%%%%%%%
\newpage
\appendix
\onecolumn

\section{CAD-Editor Dataset Preprocessing}
The DeepCAD dataset exhibits a severe data imbalance. Among deduplicated 137,012 training models, over 91.1\% have three or fewer SE operations, while only 6.2\% have 4–5 SEs. Models with more than 5 SEs make up less than 3\%, and those exceeding 10 SEs account for just 0.4\%. To mitigate the impact of this long-tail distribution, we limit our dataset to models with at most 3 SE pairs.

Additionally, the design variation generation model can sometimes introduce noise by altering a CAD model too drastically or not at all. To ensure data quality, we implement a data filtering strategy. We filter out CAD pairs with too many changes by excluding examples with more than three editing instructions or more than five \texttt{<mask>} tokens. We also filter out pairs with no significant changes by excluding instructions containing phrases like ``no transformation is needed''.

\section{Stepwise Captioning Strategy}
\label{appendix:stepwise}

\begin{figure}[ht]
\centering
\includegraphics[width=\textwidth]{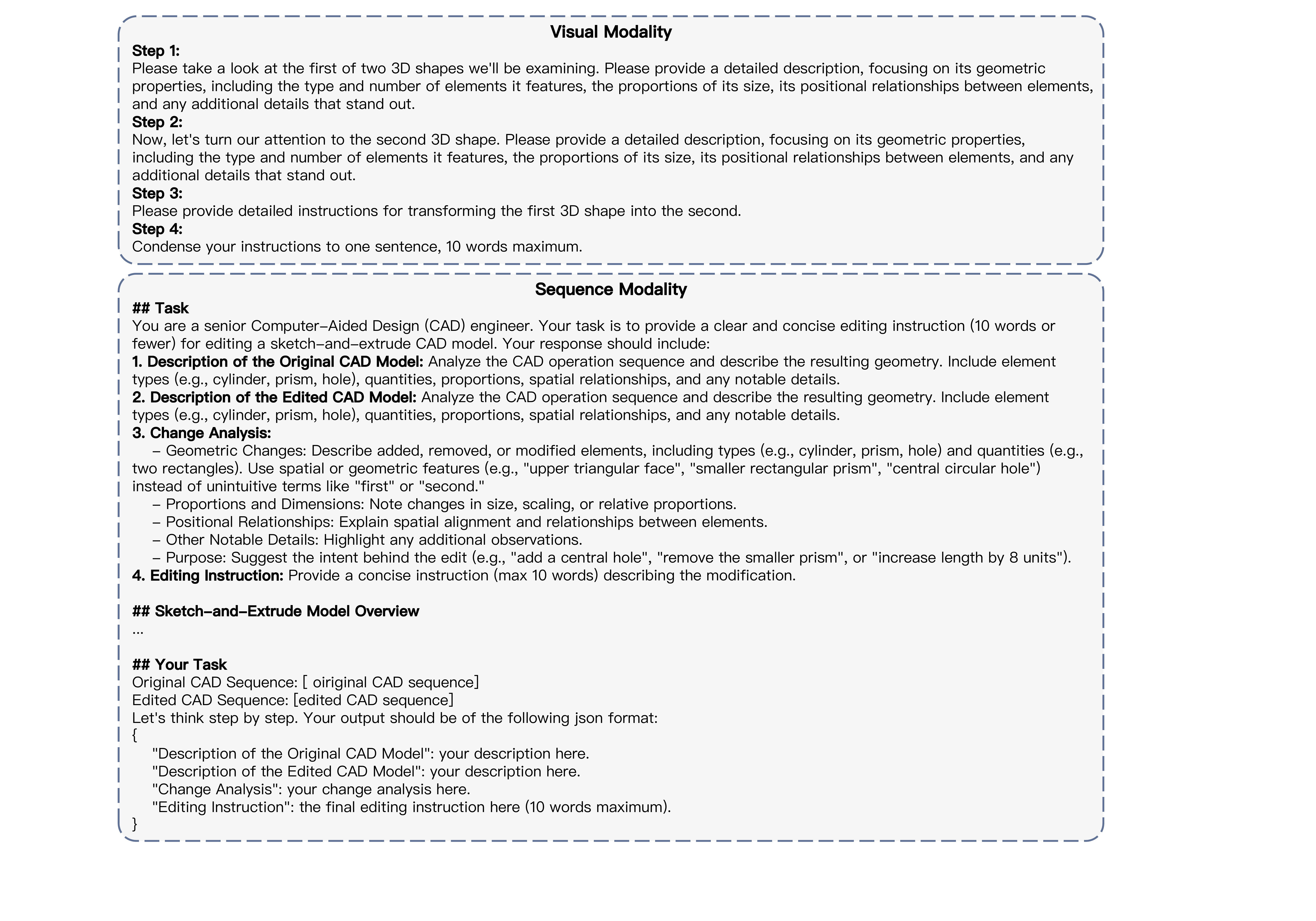}
\caption{Detailed prompt used for stepwise captioning.}
\label{fig:stepwise_captioning}
\end{figure}

To implement stepwise captioning, we utilize GPT-4o four times per CAD pair to generate the final image-level editing instructions. For sequence-level, we employ LLaMA-3-70B. The detailed prompts are illustrated in Figure~\ref{fig:stepwise_captioning}.

\begin{figure}[ht]
\centering
\includegraphics[width=0.95\textwidth]{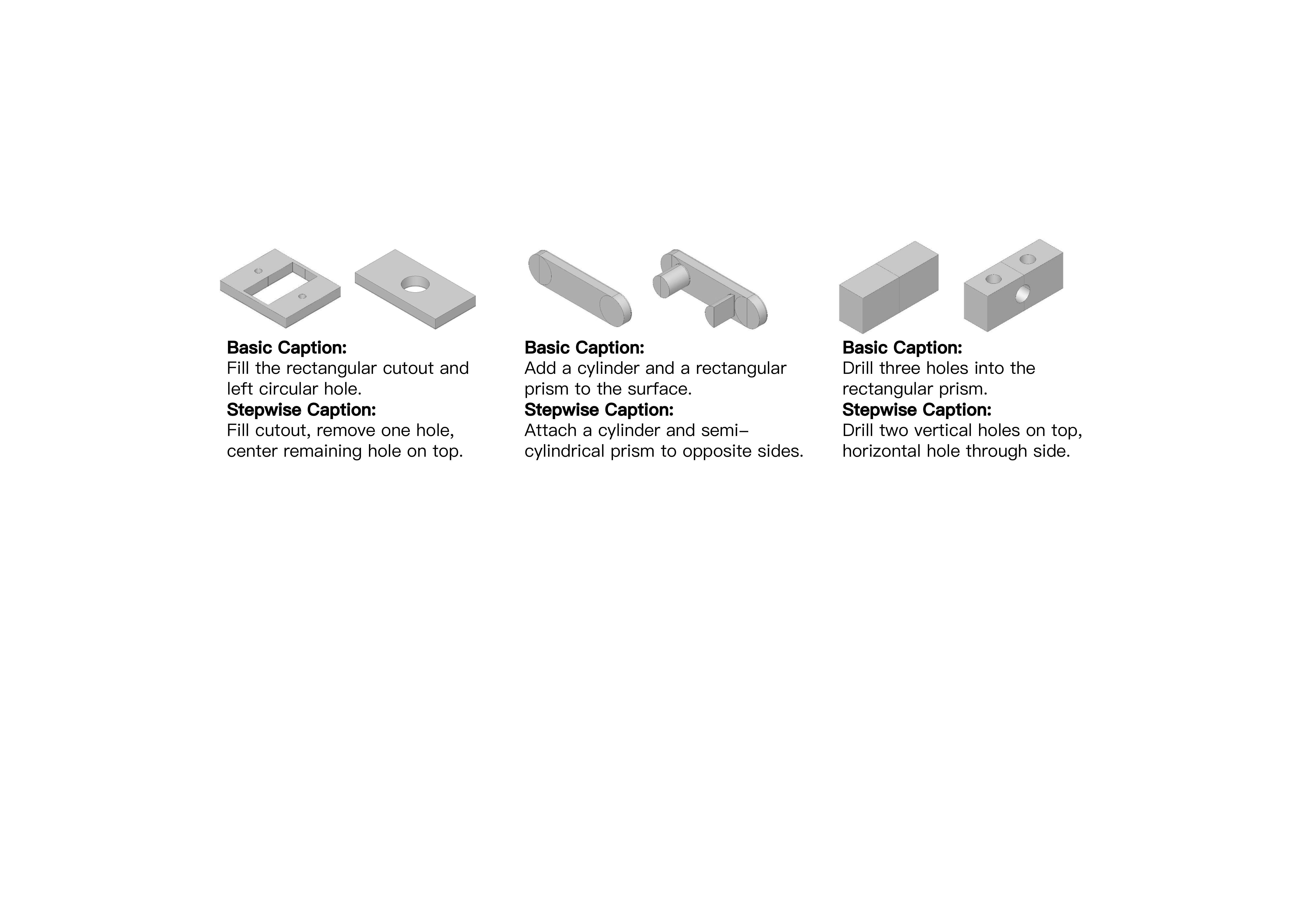}
\caption{Comparison between the basic captioning method and our stepwise captioning method.}
\label{fig:prompting_examples}
\end{figure}

Furthermore, Figure~\ref{fig:prompting_examples} presents a comparison between the basic captioning approach and our proposed stepwise captioning method.

\section{Locate-then-Infill Framework}

We adopt task-specific prompts for both the locating and infilling stages. The detailed prompt designs for each task are shown in Table~\ref{tab: task_prompt}.

\begin{table*}[h!]
\caption{Prompt for Locate-then-Infill framework.}
\label{tab: task_prompt}
\centering
\begin{tabular}{p{7cm}|p{9cm}}
\toprule
\textbf{Locating Prompt} & \textbf{ Infilling Prompt}\\ 
\midrule
 Below is a Computer-Aided Design (CAD) operation sequence. Replace the parts that need to be modified with the string ``\texttt{<mask>}" according to the editing instruction.

Original CAD Operation Sequence:

\textcolor{red}{[Original CAD sequence]}

Editing Instruction:

\textcolor{blue}{[Textual editing instruction]}

Masked CAD Operation Sequence:

\textcolor{red}{[Masked CAD sequence]} 
&
 Below is the original Computer-Aided Design (CAD) operation sequence.

Original CAD Operation Sequence:

\textcolor{red}{[Original CAD sequence]}

The parts that need to be modified according to the editing instruction have been replaced by the string ``\texttt{<mask>}".

Editing Instruction:

\textcolor{blue}{[Textual editing instruction]}

Masked CAD Operation Sequence:

\textcolor{red}{[CAD sequence with ``\texttt{<mask>}" ]}

Generate the edited CAD sequence that could replace ``\texttt{<mask>}'' in the CAD model: \\

\bottomrule
\end{tabular}
\end{table*}

\section{Additional Qualitative Results}
\begin{figure}[h]
\begin{center}
%\framebox[4.0in]{$\;$}
\includegraphics[width=\textwidth]{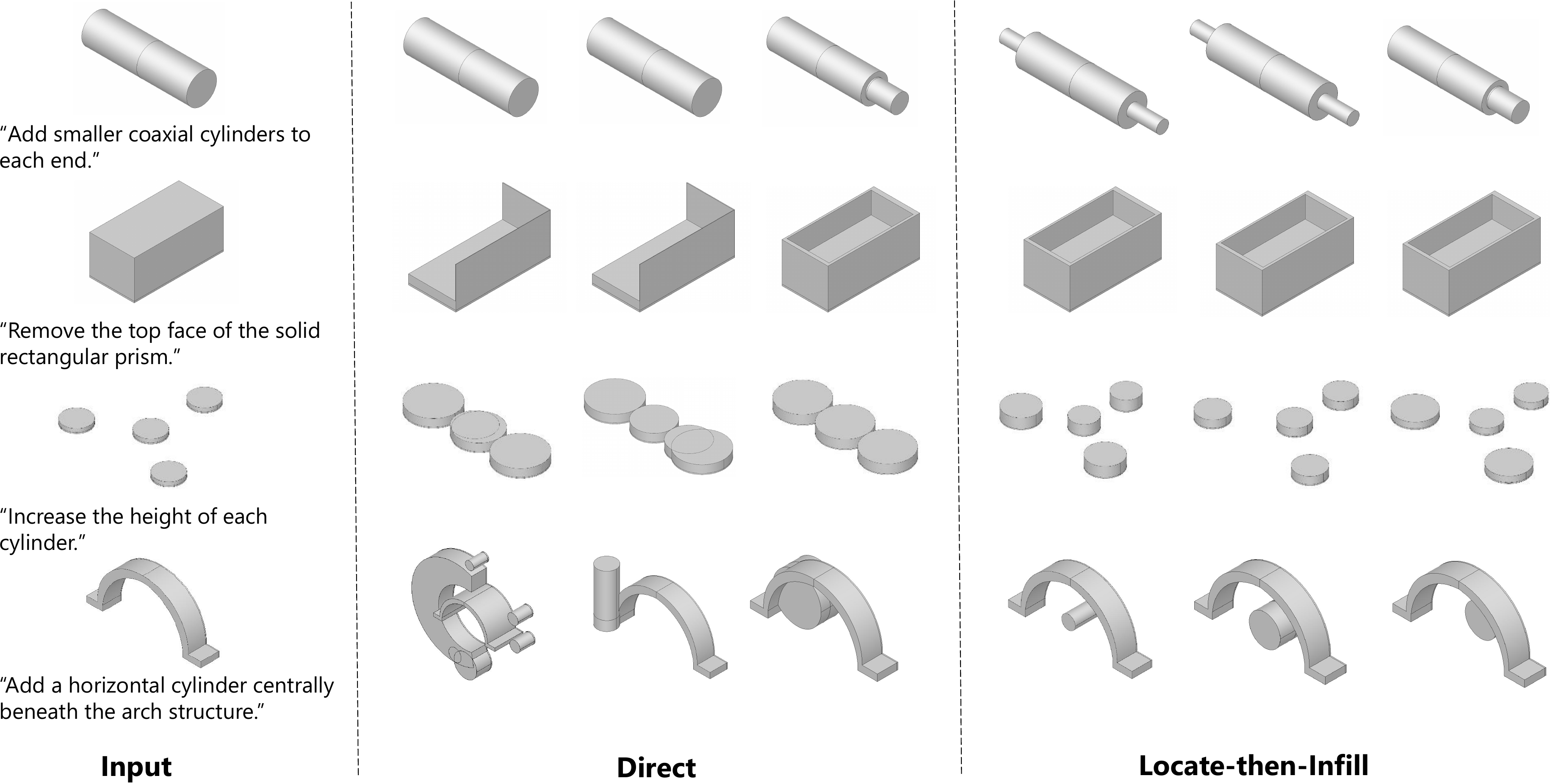}  
\end{center}
\caption{The qualitative comparison between the directly fine-tuned LLM and our Locate-then-Infill framework.}
\label{fig: comparison_locate}
\end{figure}

We present qualitative comparisons between the directly fine-tuned LLM and our proposed Locate-then-Infill framework in Figure~\ref{fig: comparison_locate}. Compared to the direct fine-tuning approach, our framework improves generation quality, enhances text-CAD alignment, and ensures greater output stability.

\begin{figure}[h]
\begin{center}
\includegraphics[width=\textwidth]{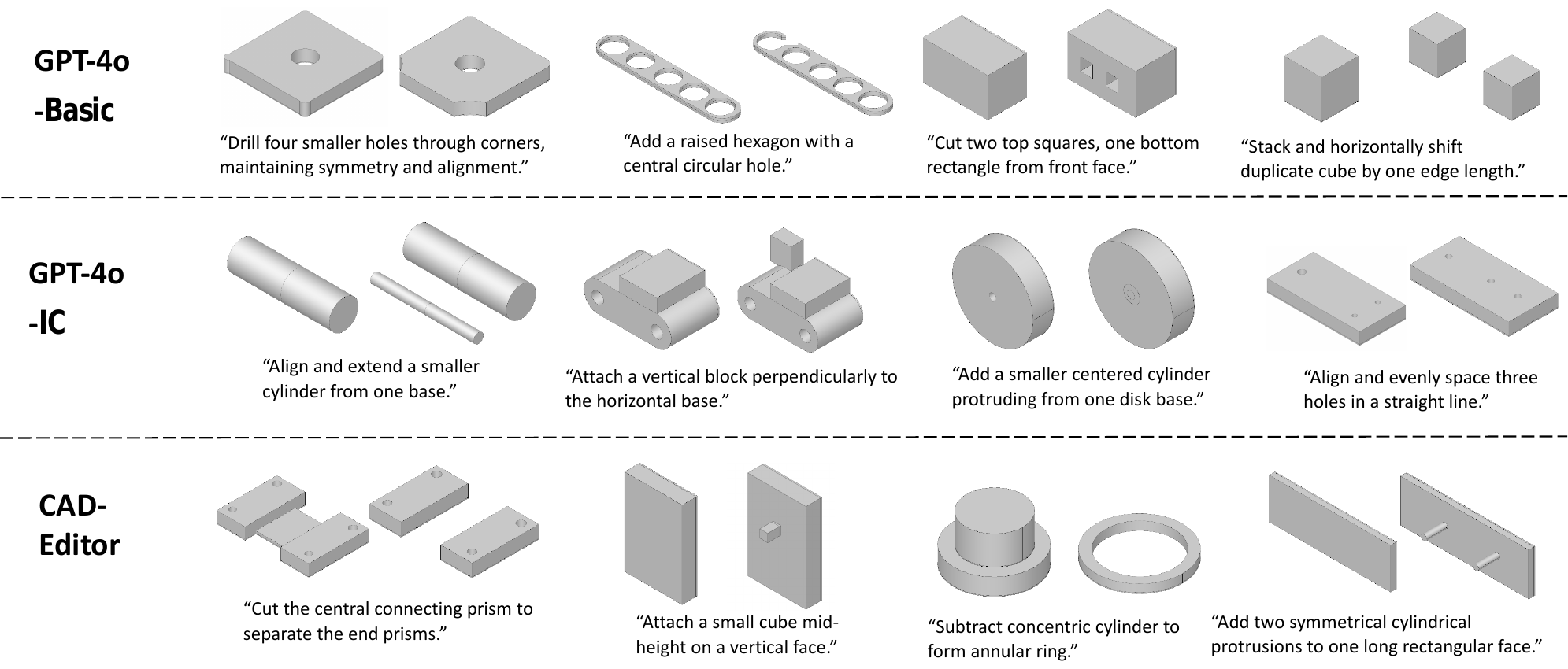}  
\end{center}
\caption{Additional qualitative comparison results between CAD-Editor and baseline methods.}
\label{fig: comparison_baselines}
\end{figure}

Additional qualitative comparisons between CAD-Editor and other baseline methods are shown in Figure~\ref{fig: comparison_baselines}, demonstrating the superior performance of CAD-Editor under various editing conditions.


\clearpage

\section{Scaling \& Generalization}
\label{Scaling & Generalization}
Due to the severe data imbalance in DeepCAD, we constrain the SE length during training. To evaluate the scalability of our method, we conduct an additional experiment in which we artificially balance the dataset across different SE lengths. Specifically, we construct a training set containing 4,000 samples for each SE length from 1 to 5, ensuring a uniform distribution. The model is then evaluated on separate test sets corresponding to each SE length.
As shown in Table~\ref{tab:scaling}, CAD-Editor consistently outperforms the baselines across all SE lengths.
Model performance declines slightly as SE length increases. Since generating longer sequences is inherently more challenging, this result demonstrates that our method generalizes well to complex CAD structures given sufficient training data.

To further assess generalization, we evaluate CAD-Editor’s generalization by testing a model trained on DeepCAD directly on Fusion 360 dataset~\cite{willis2021fusion}. As shown in Table \ref{tab:generalization} , CAD-Editor outperforms baselines, confirming its generalization ability to datasets with different shape distributions.

\begin{figure*}[h]
\begin{center}
%\framebox[4.0in]{$\;$}
\includegraphics[width=\textwidth]{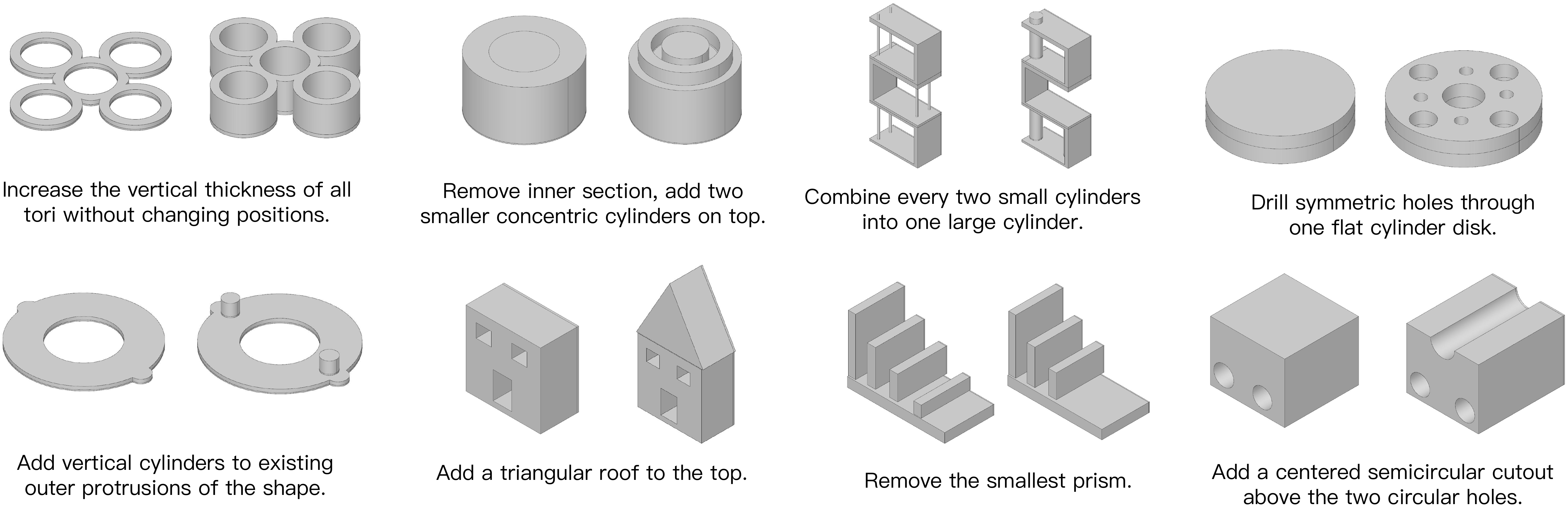}  
\end{center}
\vspace{-0.5em} 
\caption{
Extended results from CAD-Editor on more complex scenarios, including examples with more than three SEs and examples sourced from Fusion 360 datasets.
}
\label{fig:complex}
\end{figure*}

We present additional results generated by CAD-Editor in Figure~\ref{fig:complex}, including cases with more than three SEs and examples sourced from the Fusion 360 dataset.
\vspace{-0.6em} 
\begin{table}[h]
\caption{
Quantitative evaluation across different numbers of sketch-extrude (SE) operations. We construct a dataset of 20,000 examples with a uniform distribution over SE counts from 1 to 5. JSD and D-CLIP values are scaled by $10^2$. $\uparrow$: higher is better, $\downarrow$: lower is better. }
\label{tab:scaling}
\begin{small}
\begin{center}  
\setlength{\tabcolsep}{3pt}
\begin{tabular*}{\textwidth}{@{\extracolsep{\fill}}lccccccccccccccc}
\toprule
\textbf{Method} 
& \multicolumn{5}{c}{\textbf{JSD $\downarrow$}} 
& \multicolumn{5}{c}{\textbf{D-CLIP $\uparrow$}} 
& \multicolumn{5}{c}{\textbf{VR $\uparrow$}} \\
% & \multicolumn{5}{c|}{\textit{SE Number}} 
% & \multicolumn{5}{c|}{\textit{SE Number}} 
% & \multicolumn{5}{c}{\textit{SE Number}} \\
\cmidrule(lr){2-6} \cmidrule(lr){7-11} \cmidrule(lr){12-16}

& 1 & 2 & 3 & 4 & 5 
& 1 & 2 & 3 & 4 & 5 
& 1 & 2 & 3 & 4 & 5 \\
\midrule

GPT-4o-Basic
& 5.98 & 2.86 & 3.68 & 3.82 & 3.95
& -0.31 & -0.90 & -0.61 & -0.47 & -0.97
& 56.5 & 64.2 & 59.4 & 61.5 & 58.3 \\

GPT-4o-IC
& 5.42 & 1.75 & 2.29 & 2.40 & 3.67
& 0.39 & -0.06 & -0.50 & -0.11 &  -0.38
& 76.2 & 85.0 & 76.5 & 75.5 & 73.1 \\

CAD-Editor 
& 4.08 & 1.46 & 2.01 & 2.35 & 3.15 
& 0.41 & -0.23 & -0.06 & -0.02 & -0.29
& 84.3 & 91.6 & 87.5 & 79.3 & 79.5 \\

\bottomrule
\end{tabular*}
\end{center}
\end{small}
\end{table}

\vspace{-1.5em} 

\begin{table}[h]
\centering
\caption{Quantitative evaluation on the dataset generated from the Fusion 360 Gallery. Lower JSD and higher D-CLIP and VR values indicate better performance. The results demonstrate that our method exhibits strong generalization ability.}
\label{tab:generalization}
\begin{center}
\begin{tabular*}{\textwidth}{@{\extracolsep{\fill}}lccc}
\toprule
\textbf{Method} & \textbf{JSD$\downarrow$} & \textbf{D-CLIP$\uparrow$} & \textbf{VR$\uparrow$} \\
\midrule
GPT4o-Basic & 4.58 & -0.45 & 57.7 \\
GPT4o-IC    & 2.59 & -0.39 & 81.5 \\
CAD-Editor  & 2.34 &  0.41 & 93.9 \\
\bottomrule
\end{tabular*}
\end{center}
\end{table}
\vspace{-0.5em} 

\clearpage
\section{Baseline Editing Methods}
\label{appendix: baseline}
\begin{figure}[h]
\begin{center}
\includegraphics[width=\textwidth]{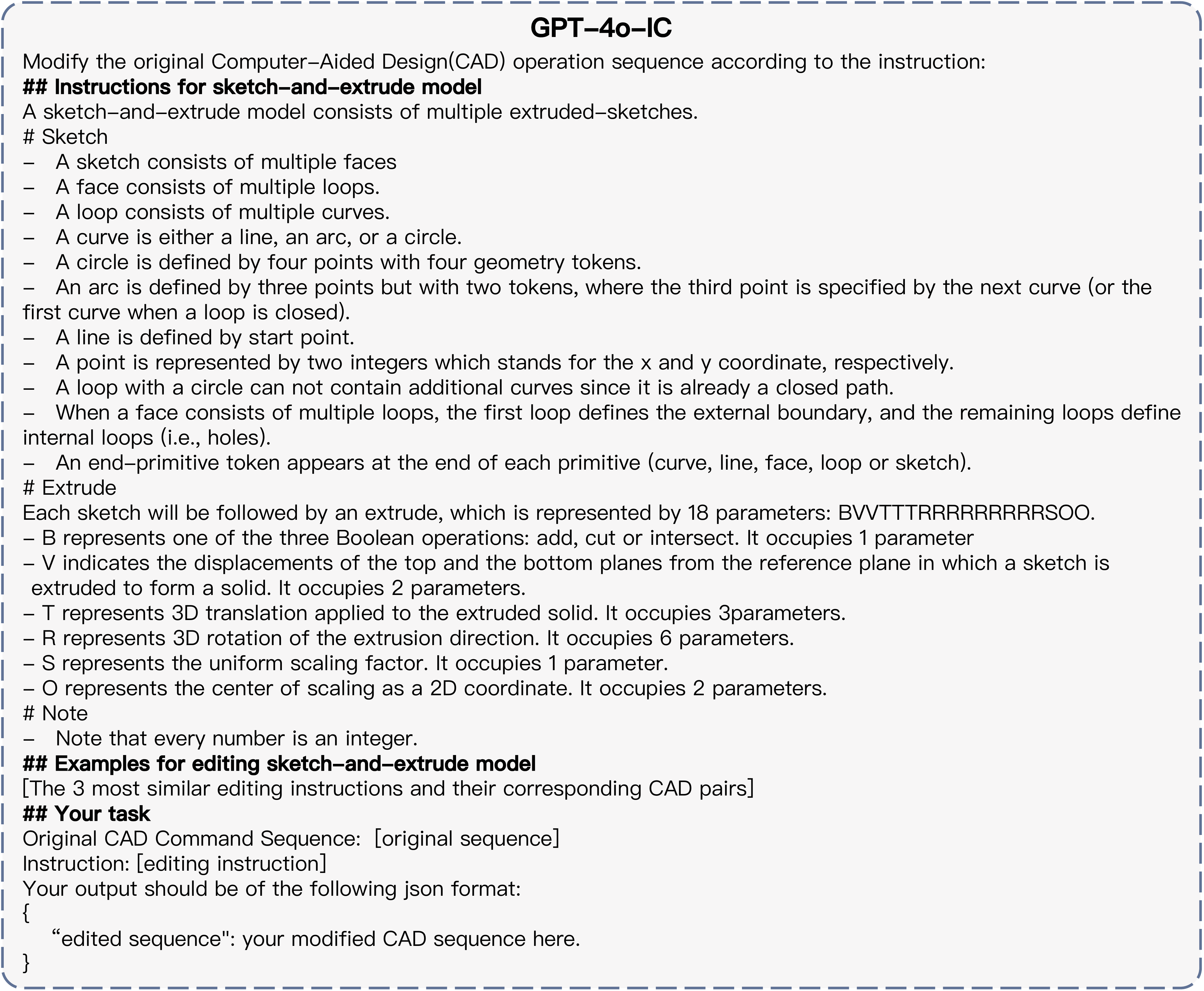}
\end{center}
\caption{Prompt employed in baseline methods for CAD model editing.}
\label{baseline_prompt}
\end{figure}

This section presents the detailed prompt used in the GPT-4o-IC baseline, as illustrated in Figure~\ref{baseline_prompt}. The GPT-4o-Basic prompt follows the same format, excluding the in-context examples.

%%%%%%%%%%%%%%%%%%%%%%%%%%%%%%%%%%%%%%%%%%%%%%%%%%%%%%%%%%%%%%%%%%%%%%%%%%%%%%%
%%%%%%%%%%%%%%%%%%%%%%%%%%%%%%%%%%%%%%%%%%%%%%%%%%%%%%%%%%%%%%%%%%%%%%%%%%%%%%%

% This document was modified from the file originally made available by
% Pat Langley and Andrea Danyluk for ICML-2K. This version was created
% by Iain Murray in 2018, and modified by Alexandre Bouchard in
% 2019 and 2021 and by Csaba Szepesvari, Gang Niu and Sivan Sabato in 2022.
% Modified again in 2023 and 2024 by Sivan Sabato and Jonathan Scarlett.
% Previous contributors include Dan Roy, Lise Getoor and Tobias
% Scheffer, which was slightly modified from the 2010 version by
% Thorsten Joachims & Johannes Fuernkranz, slightly modified from the
% 2009 version by Kiri Wagstaff and Sam Roweis's 2008 version, which is
% slightly modified from Prasad Tadepalli's 2007 version which is a
% lightly changed version of the previous year's version by Andrew
% Moore, which was in turn edited from those of Kristian Kersting and
% Codrina Lauth. Alex Smola contributed to the algorithmic style files.
\end{document}